\documentclass[conference]{IEEEtran}
\usepackage{cite}
\usepackage{xcolor}
\usepackage{graphicx}

\usepackage{tikz}

\graphicspath{{images/}{../images/}}
\usepackage[colorinlistoftodos]{todonotes}

\begin{document}
\title{Deep learning for dehazing: Comparison and analysis}

\author{\IEEEauthorblockN{Leonel Cuevas Valeriano and Jean-Baptiste Thomas}
\IEEEauthorblockA{The Norwegian Colour and Visual Computing Laboratory\\
NTNU, Gj{\o}vik, Norway\\
Email: jean.b.thomas@ntnu.no}\\
\IEEEauthorblockN{Alexandre Benoit}
\IEEEauthorblockA{LISTIC\\
Univ. Savoie Mont Blanc, Annecy, France}}

\maketitle

\begin{abstract}
    We compare a recent dehazing method based on deep learning, Dehazenet, with traditional state-of-the-art approaches, on benchmark data with reference. Dehazenet estimates the depth map from transmission factor on a single color image, which is used to inverse the Koschmieder model of imaging in the presence of haze. In this sense, the solution is still attached to the Koschmieder model. We demonstrate that the transmission is very well estimated by the network, but also that this method exhibits the same limitation than others due to the use of the same imaging model.
\end{abstract}

\section{Introduction}
\label{sec:intro}

Dehazing aims at improving visibility in images captured in a presence of haze. In general, methods can be classified in two categories (Jessica El Khoury PhD~\cite{jess2016}, Chapter 3). One is based on an image enhancement paradigm, and is usually instances of local histogram equalization. The other category aims at the inversion of the Koschmieder model~\cite{harald1924theorie} written in Equation~\ref{eq:koschmieder}, 
\begin{equation}
J(x)=I(x)t(x)+A_\infty(1-t(x)),
\label{eq:koschmieder}
\end{equation}
which states that the image captured at a position $x$ yields $J(x)$ that is a linear combination of the radiant image $I(x)$ and the contribution of the airlight $A_\infty$ weighted by a transmission factor $t(x)$. $A_\infty$ is defined as the light sent to the camera by an object at infinite distance, {\itshape i.e.} the diffusion of the light by the haze. The transmission factor is $t(x)=\exp(-\beta d(x))$, where $\beta$ is the scattering coefficient of the haze and $d(x)$ the distance of the object from the camera. This is performed by estimating $t(x)$ and $A_\infty$ separately or jointly.

Recently, solutions using deep learning have been introduced. They fall into this category. For instance Dehazenet \cite{cai2016dehazenet} focuses on the estimation of t(x), and AOD-Net \cite{Li_2017_ICCV} focuses of the joint estimation of $t(x)$ and $A_\infty$. Two limitations can be observed: 1-the Koschmieder model seems to have a limited validity for heavy amount of haze where the airlight contribution is predominant compared to the radiant signal (Jessica El Khoury PhD~\cite{jess2016}), 2-Networks are trained on simulated material based on the Koschmieder model, which impacts generalization capabilities and thus may limit performance levels on real data. We perform a quantitative analysis of the first point and compare Dehazenet with the state of the art on the CHIC database \cite{chicICIP, ElKhoury2018}. The methods selected for comparison are: DCP \cite{he2011single}, FAST \cite{tarel2009fast} and CLAHE \cite{xu2009fog}; DCP is based on the inversion of the Koschmieder model by use of the Dark Channel to estimate $t(x)$, FAST proposes a variation of this by estimating an \textit{Athmospheric Veil} which represents the achromatic veil responsible for intensity changes in the image due to the light interactions in the environment, and finally CLAHE is based on applying contrast-limited adaptive histogram equalization.

The remaining of the paper is organized as follows. Next Section considers the description of the parameters estimation by learning, Section~\ref{sec:exp} defines the experimental procedure. Benchmarking results are presented in Section~\ref{sec:res} and demonstrate that dehazenet exhibits the same limitations than other state-of-the-art methods based on the Koschmieder model despite a very good estimation of the transmission factor.

\section{Deep learning for dehazing}
\label{sec:dl}

Researchers have recently turned their attention to deep learning in order to explore how well it performs in the task of haze removal, inspired by the outstanding results of Convolutional Neural Networks (CNN) in high-level vision tasks such as image classification~\cite{krizhevsky2012imagenet}. The main reasoning behind these methods is the fact that the human brain can quickly identify the hazy area from the natural scenery without any additional information~\cite{cai2016dehazenet}, so by the use of CNNs it seems plausible to extract the features necessary to perform the dehazing task. Although more efficient or vision inspired formulation may rise in the future, the first attempts focus on the inversion of the Koschmieder imaging model.

\textbf{DehazeNet:}
Cai {\itshape et al.}~\cite{cai2016dehazenet} introduced DehazeNet, a deep learning  method for single image haze removal. DehazeNet is, because of its principles, a method based on the inversion of the Koschmieder model. It proposes a new approach to estimate the transmission map $t(x)$. Given a hazy image as input, DehazeNet computes and output its corresponding $t(x)$, which can be then used to recover a haze-free image by inversion of the model. DehazeNet is trained with thousands of hazy image patches, which are synthesized from haze-free versions of images taken from the Internet. Since the model for generating these hazy patches is known, a ground truth for the transmission map $t(x)$ can be provided for training. DehazeNet only estimates $t(x)$, the estimation of the global atmospheric light $A_\infty$ is done in a separate and independent step. In fact it can be done by using any of the approaches used in other methods. For the purpose of this study, we have used the same $A_\infty$ estimation for all methods in the benchmark. 

\textbf{AOD-Net:}
More recently Li {\itshape et al.} introduced the All-in-One Dehazing Network (AOD-Net)~\cite{Li_2017_ICCV}. Unlike DehazeNet, AOD-Net is based on a reformulation of the Koschmieder model, where all parameters are encapsulated into one variable. This enables the joint implicit estimation of the transmission map and the atmospheric light to invert the model. For the training of AOD-Net, the authors created synthesized hazy images, using the ground-truth images with depth metadata from the indoor NYU2 Depth Database \cite{silberman2012indoor}, which contains around 30,000 images. Different atmospheric light and $\beta$ values are set to generate a new set of hazy images which are fed as input to AOD-Net. The output of the network is then compared to the original haze-free image and trained to minimize the error between the output and the original images prior haze simulation. They also used a set of natural hazy images to evaluate visually the general performance. Since the AOD-Net has been published after our experiment, we developed the analysis around Dehazenet in this paper. AOD-Net will be investigated in further work.

\section{Experiment}
\label{sec:exp}

The objective of our work is to perform an objective comparison between deep learning based methods such as DehazeNet and traditional haze removal methods. We opted for different approaches to perform the comparison. We made use of the CHIC database introduced by El Khoury {\itshape et al.} \cite{chicICIP, ElKhoury2018}. CHIC stands for "Color Hazy Image for Comparison". The database consists of two different scenes that include numerous objects with different shapes, colors, positions, surface types (glossy or rough surfaces) and textures. The scenes also include four Macbeth color checkers at different distances for color accuracy checks. A set of high-resolution images of the same scene is acquired under a controlled environment with the same light conditions; first without any haze (haze-free image) and then under 9 different levels of fog density which is introduced by using a fog machine. Each image has a uniform fog level. Level 9 is the lowest fog density while level 1 is the highest. In addition to this, for each scene the distances of the four color checkers to the camera are known, the fog properties, such as the scattering coefficient ($\beta$) are also known and the lighting conditions remain constant.

\begin{figure}[htb]
    \centering
    \includegraphics[width=8cm]{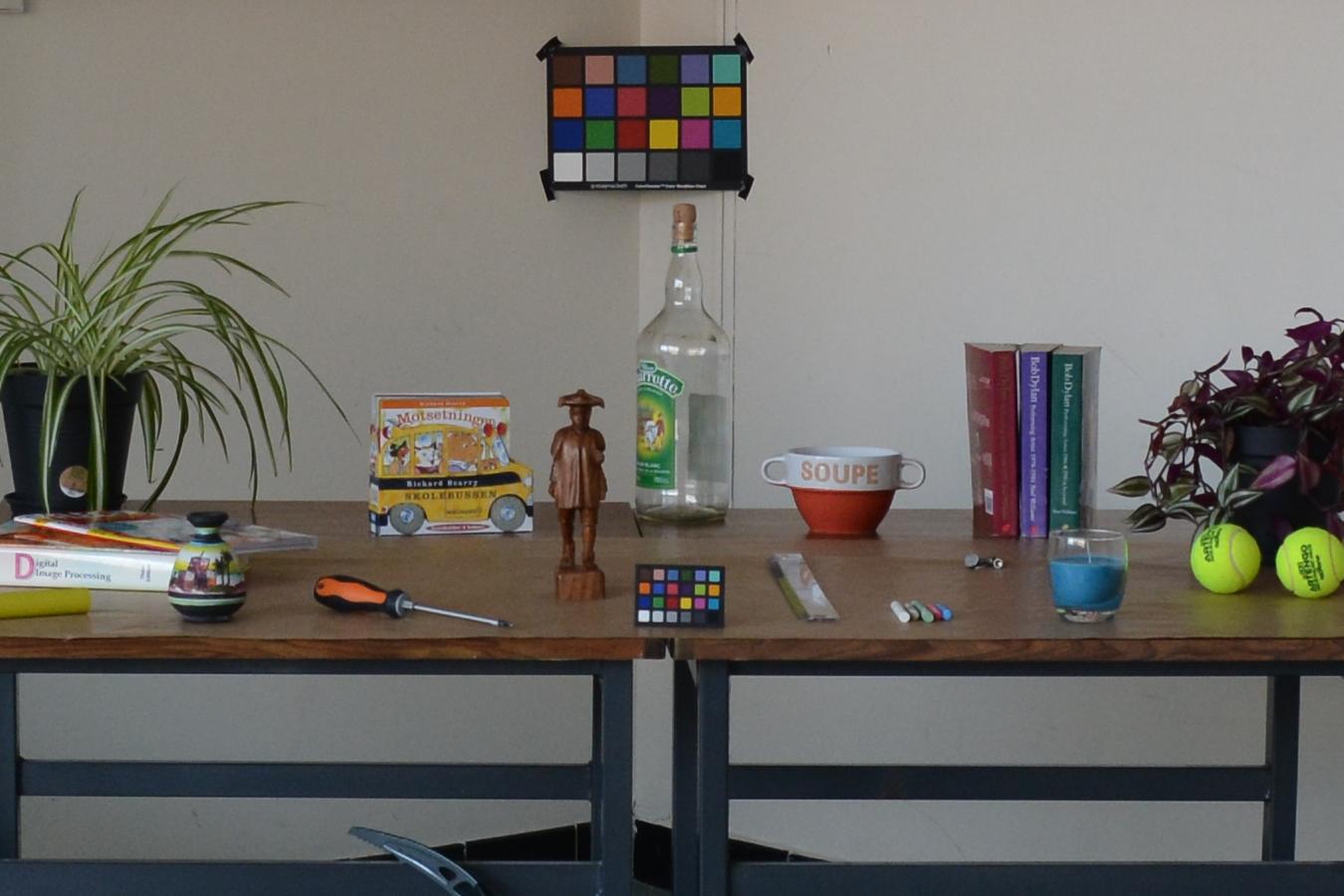}
    \caption{Selected regions of the scene to do the comparisons.}
    \label{fig:chosen_scene}
\end{figure}

Due to the high resolution of the images and because of the limitations in our computing power it was not possible to give the full image as an input to DehazeNet, to overcome this issue we decided to crop a region of the original image and use it as input for DehazeNet and the other reference methods. We chose this option because we were concerned about the possibility that resizing processes could introduce artifacts in the image, which might have an impact on the Koschmieder model and the results. The particular region of the scene of the CHIC database shown in Figure~\ref{fig:chosen_scene}. We selected this region of the scene due to the presence of the color checker in the back of the scenes, which will be used for comparison. Also it covers objects at different depths and being the furthest (with more depth) part of the scene it proves to be a more challenging task for the different haze removal methods.

The $A_\infty$ is measured on the dataset and given to all the algorithms, which enable a fair comparison.

A set of state-of-the-art dehazing methods is considered for comparison for different levels of haze : DCP \cite{he2011single}, FAST \cite{tarel2009fast}, CLAHE \cite{xu2009fog} and Dehazenet. For this, the pre-trained Caffe implementation provided by Zlinker \cite{zlinker_git} is used.

We compute a selection of metrics for dehazing evaluation \cite{ElKhoury2017}, which mostly indicates how well edges are recovered or enhanced ($e$\cite{hautiere2011blind}, $r$\cite{hautiere2011blind}, FADE \cite{choi2015referenceless} and VSI \cite{zhang2014vsi}). We perform a color analysis of the results based on a color checker, which indicates how well color are recovered (similar to \cite{elkhoury:hal-01202989}).
We also investigate how well DehazeNet estimates the transmission factor $t(x)$.

\section{Evaluating DehazeNet}
\label{sec:res}

\subsection{Transmission map estimation analysis}

DehazeNet performs an estimation of the transmission map $t(x)$. We propose to evaluate this directly compared to measured values from the benchmark. We also compare with the transmission map obtained by DCP under different levels of haze. We only compared with DCP because FAST is based on a different variation of the model and CLAHE does not try to estimate the transmission map.

The distance between the color checkers and the camera serves as a ground truth to compare the two considered methods. Since an approximation of the scattering coefficient $\beta$ is available for each scene, a ground truth $t(x)=\exp(-\beta d(x))$ can be computed for each of those color checkers to benchmark the considered methods. 

\begin{figure}[htb]
    \centering
    \includegraphics[width=8cm]{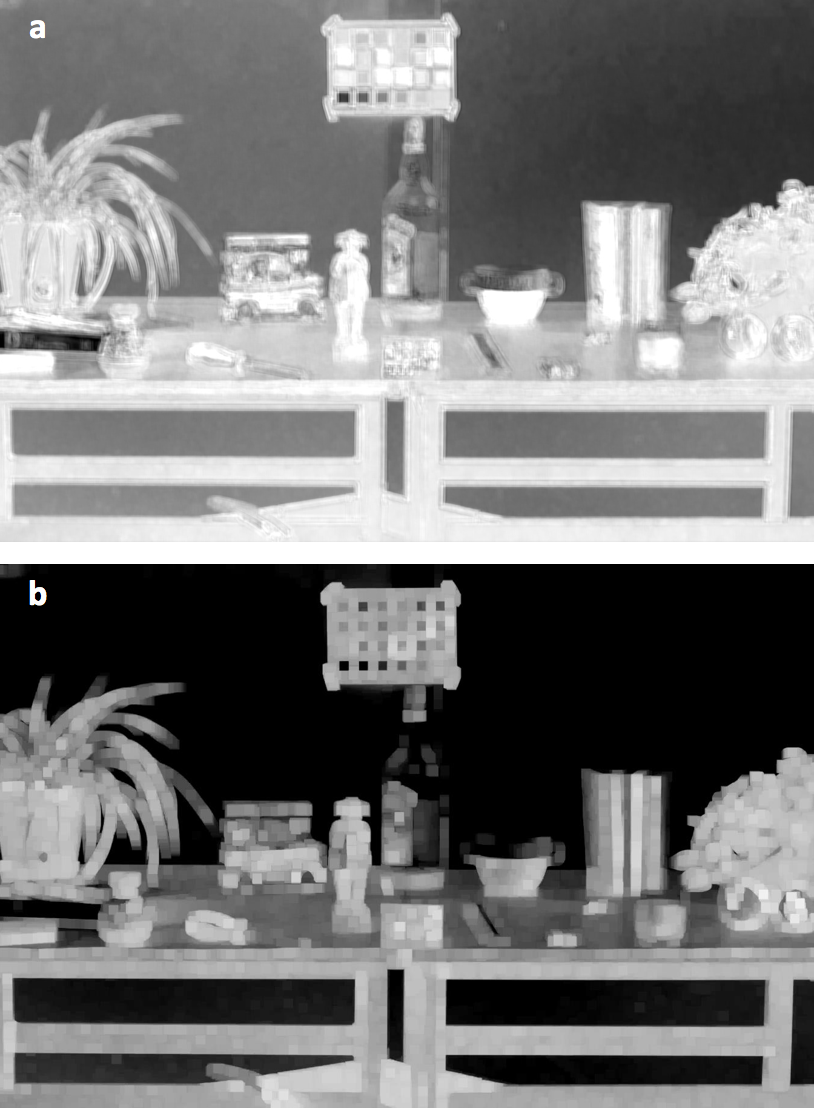}
    \caption{Examples of transmission maps. a) Transmission map estimated by Dehazenet for haze level 9. b) Transmission map estimated by DCP for haze level 9. }
    \label{fig:tmaps}
\end{figure}

We decided to perform the quantitative comparison only for level 5 and above, where the model inversion permits to improve the visibility of the scene. Levels below 5 destroy so much the scene that the transmission map, as formulated, is not relevant. Figure~\ref{fig:tmaps} shows an example of the transmission maps obtained by DehazeNet and DCP for level 9. Already visually, we could spot differences.

In order to get an estimation of the transmission of the whole color checker we segment it out and take all the pixels in the region, we exclude the highest and lowest 15\% values, then calculate the mean of the remaining 70\%. We do this in order to avoid the effect of noise and outliers due to bad estimations. The obtained estimation is then compared to the values obtained using the ground truth data for the scene (color checker on the back is located 7 m from the camera and the one on the table at 4.35 m from the camera). Tables~\ref{tmap_res_cc1} and~\ref{tmap_res_cc2} show the scattering coefficients ($\beta$) used, the transmission values obtained from both DehazeNet and DCP, along with the expected values calculated using the ground truth data, for the color checker on the back and the color checker of the table, respectively. We observe that DehazeNet clearly outperforms DCP with results fairly close to the measured ground truth. This is especially true for stronger haze.

\begin{table}[h!]
\footnotesize
\centering
\caption{Results of transmission estimation analysis for the Color Checker placed on the back of the scene.}
\label{tmap_res_cc1}
\begin{tabular}{ccccc}
{\color[HTML]{000000} \textbf{\begin{tabular}[c]{@{}c@{}}Haze\\ Level\end{tabular}}} & {\color[HTML]{000000} \textbf{$\beta$}} & {\color[HTML]{000000} \textbf{\begin{tabular}[c]{@{}c@{}}DehazeNet\end{tabular}}} & {\color[HTML]{000000} \textbf{\begin{tabular}[c]{@{}c@{}}DCP\end{tabular}}} & {\color[HTML]{000000} \textbf{\begin{tabular}[c]{@{}c@{}}Ground\\ truth\end{tabular}}} \\
5 & 103.69 & 0.559 & 0.133 & 0.484 \\
7 & 83.57 & 0.618 & 0.255 & 0.557 \\
9 & 17.84 & 0.855 & 0.617 & 0.883
\end{tabular}
\end{table}

\begin{table}[ht]
\footnotesize
\centering
\caption{Results of transmission estimation analysis for the Color Checker placed on the table on the scene.}
\label{tmap_res_cc2}
\begin{tabular}{ccccc}
{\color[HTML]{000000} \textbf{\begin{tabular}[c]{@{}c@{}}Haze\\ Level\end{tabular}}} & {\color[HTML]{000000} \textbf{$\beta$}} & {\color[HTML]{000000} \textbf{\begin{tabular}[c]{@{}c@{}}DehazeNet\end{tabular}}} & {\color[HTML]{000000} \textbf{\begin{tabular}[c]{@{}c@{}}DCP\end{tabular}}} & {\color[HTML]{000000} \textbf{\begin{tabular}[c]{@{}c@{}}Ground\\ truth\end{tabular}}} \\
5 & 103.69 & 0.624 & 0.262 & 0.637 \\
7 & 83.57 & 0.700 & 0.432 & 0.695 \\
9 & 17.84 & 0.872 & 0.725 & 0.925
\end{tabular}
\end{table}

\subsection{Color accuracy}

Several haze-removal methods are known to make colors significantly more colorful, and most methods disregard the color accuracy of the restored images. In addition, the color reliability is an aspect that most visibility metrics do not take into account. We propose to investigate on the color accuracy of the restored images by making use of the Macbeth Color Checkers in the scene and of the haze-free reference images, which are provided in the CHIC database. 

\begin{figure}[htb]
    \centering
    \includegraphics[width=8cm]{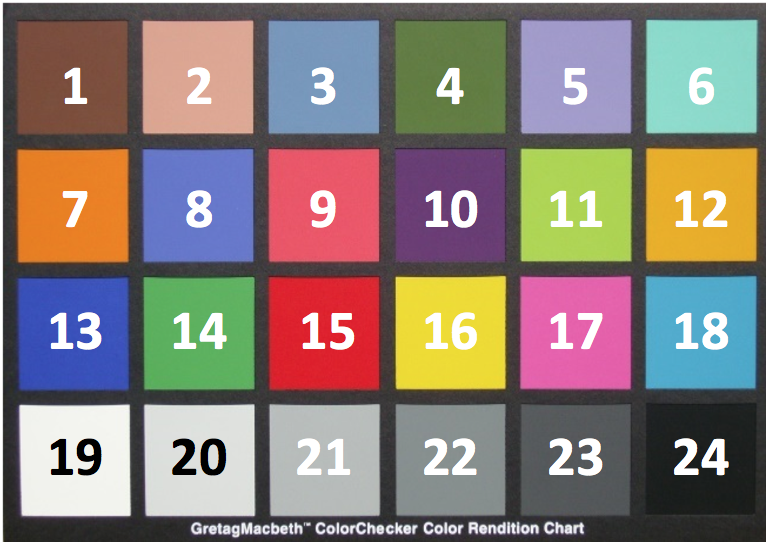}
    \caption{Labels for the 24 GretagMacbeth ColorChecker patches.}
    \label{fig:color_checker}
\end{figure}

We select several patches of the color checker and plot on a {\itshape rg} chromaticity diagram the color for the hazy image, the haze-free image, and the color restored by DehazeNet and the benchmarking algorithms. This allows us to make a comparison of the color between the different methods. In Figure \ref{fig:color_checker}, we present the labeled patches. We only show the results for low densities of haze because with level 5 and lower, the restored colors have values close to the ones of the achromatic patches. For those levels, visualization is meaningless. In practice, we select 8 patches (2, 6, 7, 13, 14, 16, 17, 19). These patches were chosen to cover different regions of the space. Only one of the achromatic patches were selected because their results are very similar.

\begin{figure}[htb]
    \centering
    \includegraphics[width=9cm]{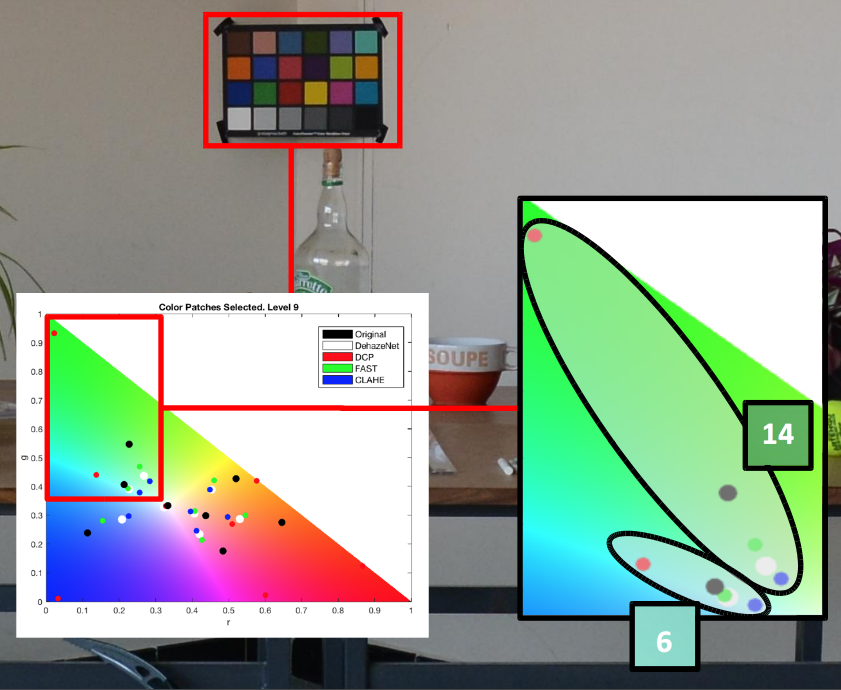}
    \caption{Results example of  for haze level 9. We can observe a similar trend for the results with DCP (in red) showing a significant over-estimation of colorfulness.}
    \label{fig:cc_example}
\end{figure}

We present an example of the results and its interpretation in Figure~\ref{fig:cc_example}. We can observe that for level 9, the restored colors are  close to those of the ground truth, DehazeNet comes in second place after FAST with an average distance of 0.081 units. We can also observe that DCP tends to over-enhance the results, as we will see later for other levels of haze.

\begin{figure}[htb]
    \centering
    \includegraphics[width=9.2cm]{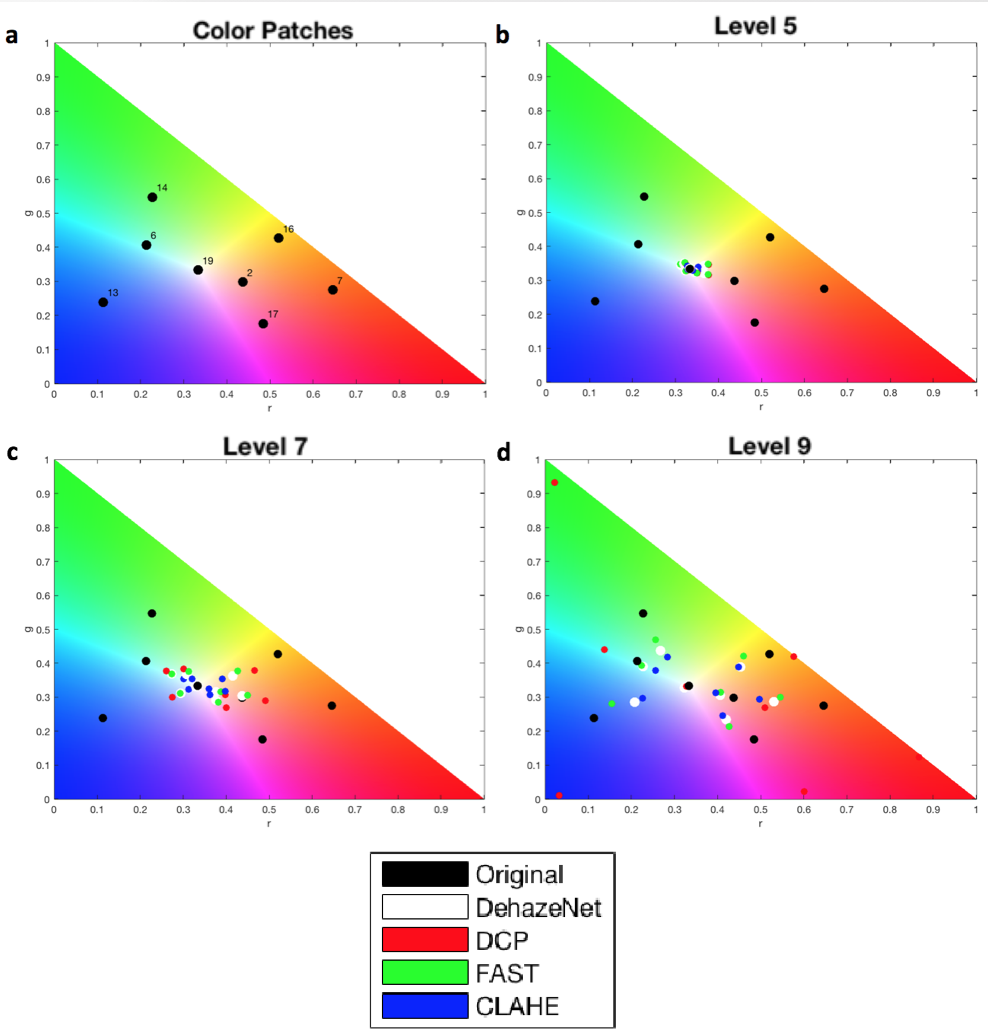}
    \caption{Image of the {\itshape rg} chromaticity for the 8 selected patches in a)the original haze-free image, b) under haze level 5, c) under haze level 7 and d) under haze level 9.}
    \label{fig:rg_results}
\end{figure}

In Figure~\ref{fig:rg_results} (b) - (c) we present, for representative levels of haze, the {\itshape rg} chromaticity of the obtained colors for the selected patches by using DehazeNet in order to compare with those obtained using DCP, FAST and CLAHE, for a complete comparison we also include the colors of the haze-free image and the hazy image. 

For other levels of haze we can observe different phenomena: for level 5 the restored colors are still mostly pale grayish, so the real colors are still quite far from the ground truth, in this case DehazeNet comes in second place, only behind FAST, with an average distance of  0.227 units. For level 5, the restored colors by all methods seem to approach closer to those of the ground truth. In this case, DehazeNet falls behind DCP and FAST with an average distance of 0.153 units. 

Overall, DehazeNet's performance in regards to color accuracy is good, giving in most cases the first or second-closest method (most times second, since FAST normally gives the closest one) to the original haze-free color and without showing significant over-enhancements, unlike DCP. This is to put in relation with the excellent estimation of the transmission map.

\subsection{Image analysis. Model limitations}

\begin{figure}[htb]
    \centering
    \includegraphics[width=8cm]{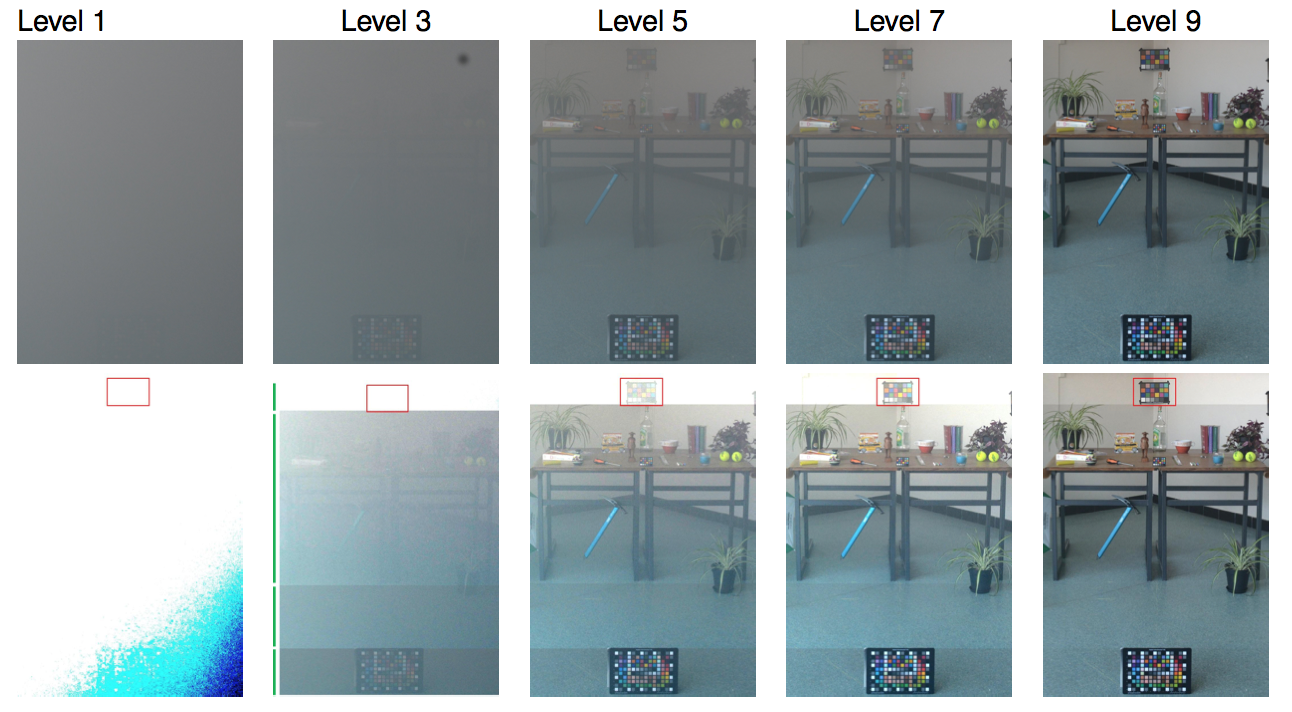}
    \caption{Evaluation of the Koschmieder model limitations under different levels of haze. The upper row shows the original images, while the bottom row shows the dehazed versions. Note that below level 5 the color checker in the back can no longer be restored. Image reproduced from \cite{jess2016}.}
    \label{fig:kosh_limit}
\end{figure}

El Khoury {\itshape et al.} \cite{jess2016} took advantage of the CHIC database to verify the assumptions and limitations of the Koschmieder model. Having the color checkers as reference points, they divided the scene into four different parts and, by using the known distances of the color checkers in the scene, they were able to calculate the transmission map for each part of the scene, which they would later use to obtain a "haze-free" version of the scene as seen in Figure~\ref{fig:kosh_limit}.

\begin{figure*}[htb]
    \centering
    \includegraphics[width=\textwidth]{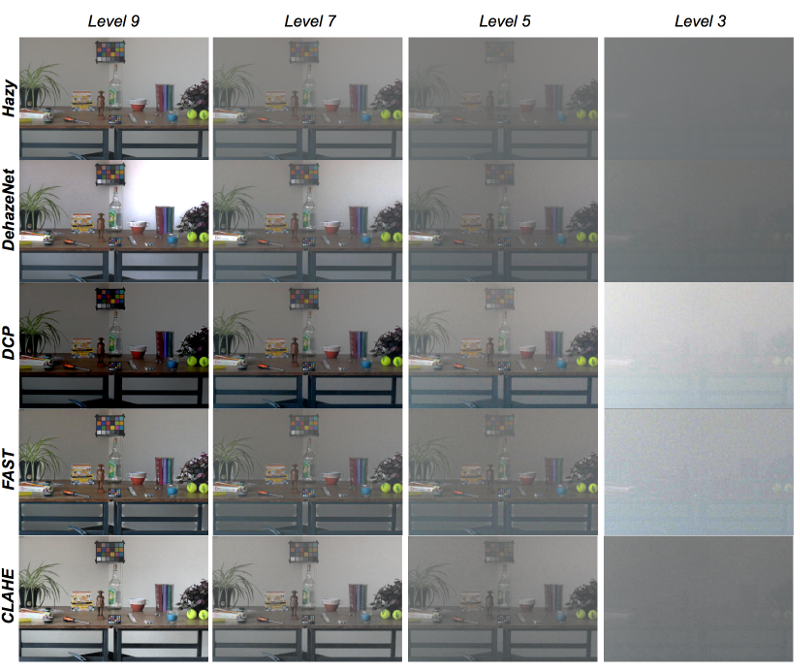}
    \caption{Results obtained by applying the selected haze removal methods on the selected cropped region of the scene under different levels of haze. From top to bottom: hazy input image, DehazeNet results, DCP results, FAST results and CLAHE results.}
    \label{fig:res_all_methods}
\end{figure*}

This, in a way, presents the limits of the inversion model itself, since the transmission is calculated using known depths instead of only an estimate. So, as a validation step, we evaluated the results of DehazeNet under different levels of haze. Our idea was to judge visually how good DehazeNet performs by having the "limit results" of the model for comparison. In Figure~\ref{fig:res_all_methods}, we show the results obtained for all the methods.

The results show an improvement in the overall quality of the image for all the different haze levels above 3. In level 3 we can still see a slight improvement of the visibility particularly in the book on the left and the tennis balls on the right side of the table. We can observe that the color checker on the back is no longer visible in the restored images of level 3 haze for any of the methods. This is coherent with the limitations of the Koschmieder model \cite{jess2016}. Although, when comparing with the results seen in Figure~\ref{fig:kosh_limit}, there is still possibility for improvement. So we concluded that the performance of DehazeNet is on par with other state-of-the-art methods, but still constrained by the limitation of the Koschmeider model. 

\subsection{Image analysis. Metrics}

The final comparison consists in using a set of metrics which are normally used for the quality assessment of visibility enhancement algorithms. For this comparison the following metrics were chosen based on the recommendations by El Khoury {\itshape et al.} \cite{ElKhoury2017}:
\begin{itemize}  
\setlength\itemsep{0.25em}
\item \textbf{\textit{e} and \textit{r}:} this metric compares the restored image to a reference hazy image. The \textit{e} index evaluates the ability of the method to restore edges which were not visible in the reference image but are in the restored image, while the \textit{r} index is an indicator of restoration quality. The scores of these indexes refer to the gain of visibility, higher score means better results obtained \cite{hautiere2011blind}.
\item \textbf{FADE:} Fog Aware Density Evaluator. This is a reference-less metric. It is based on natural scene statistics. They create a model based on extracted features observed in 500 natural hazy and 500 haze-free images. By using this model the metric estimates the "fog density" in the image, therefore lower scores represent a better restored image. It performs particularly well at assessing color recovery assessment \cite{choi2015referenceless}. 
\item \textbf{VSI:} Visual Saliency-based Index. This is one of the few metrics that compares the restored image to the original haze-free image. It is a metric based on the assumption that an image’s Visual Saliency map has a close relationship with its perceptual quality. It is based on their own Saliency Detection by combining Simple Priors (SDSP) method \cite{zhang2013sdsp} that works by integrating prior knowledge from frequency, color and location. VSI outperforms the other two metrics significantly when it comes to sharpness recovery assessment \cite{zhang2014vsi}.
\end{itemize}

\begin{figure}[htb]
    \centering
    \includegraphics[width=8cm]{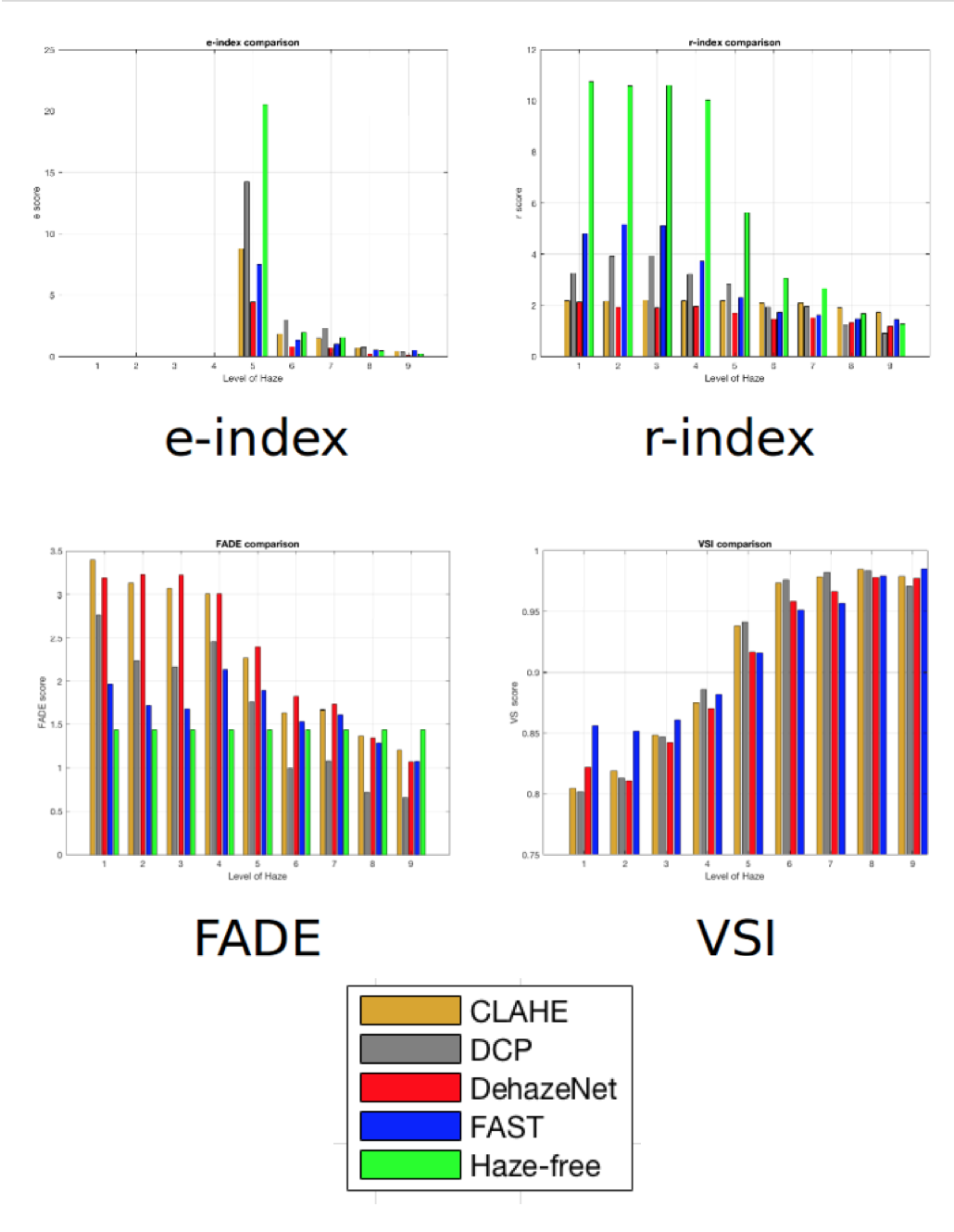}
    \caption{Results of the selected metrics under the different available haze levels. Note that the performance varies depending on the haze level, but DehazeNet (in red) tends to perform poorly across all metrics (note: for FADE values lower is better).}
    \label{fig:res_metrics}
\end{figure}

The results of using these metrics are shown in Figure~\ref{fig:res_metrics}, when appropriate, we include the results for the original hazy image as well. Note that we do not present the results of the \textit{e} index for levels under 5, because due to the low visibility conditions any change in the image results in very high values for all methods, which are difficult to compare. The results show that in general (across all the four metrics), over most of the different levels of haze, DehazeNet performs worse than all the other methods.
So, by considering the results of these metrics, we can conclude that DehazeNet performs poorly in terms of edge visibility and structural sharpness restoration, either being the worst or second-worst performer in the majority of cases. The results of those metrics should not be generalized to the overall evaluation of the methods since it has been demonstrated that none of them correlate perfectly with the visual observation on this database~\cite{ElKhoury2017}.

\section{Conclusion}
\label{sec:cln}

DehazeNet performs comparable to other state of the art algorithms. We observed better results for transmission map estimation and color accuracy, but worse for improvements in edge visibility and structural sharpness. This analysis is based only on an image across several level of one type of fog. We nevertheless can predict that eventually, deep learning will permit to recover the parameters of the model the best possible. AOD already improved a lot on the Airlight estimation, if it is re-tuned for real scenes, it may become optimal.
Those proposals still face the limitations of the imaging model, and investigations in the reformulation of the model of haze are required to create a breakthrough in performance.

\bibliographystyle{IEEEtran}
\bibliography{references}

\end{document}